\documentclass[10pt,twocolumn,letterpaper]{article}

\usepackage[pagenumbers]{cvpr} 
\usepackage{svg}
\usepackage{fancyhdr}
\usepackage{placeins}

\definecolor{cvprblue}{rgb}{0.21,0.49,0.74}
\usepackage[pagebackref,breaklinks,colorlinks,allcolors=cvprblue]{hyperref}
\usepackage{float}

\def\paperID{*****} 
\def\confName{CVPR MIV Workshop}
\def\confYear{2025}

\title{Steering CLIP's vision transformer with sparse autoencoders}

\author{
Sonia Joseph$^{1,2}$ \quad
Praneet Suresh$^{1,2}$ \quad
Ethan Goldfarb$^{3}$ \\
Lorenz Hufe$^{4}$ \quad
Yossi Gandelsman$^{5}$ \quad
Robert Graham$^{2}$ \\
Danilo Bzdok$^{1,2}$ \quad
Wojciech Samek$^{4}$ \quad
Blake Aaron Richards$^{1,2}$ \\[0.3cm]
$^{1}$Mila \quad
$^{2}$McGill University \quad
$^{3}$Independent Researcher \\
$^{4}$Fraunhofer HHI \quad
$^{5}$UC Berkeley \\[0.3cm]
\texttt{sonia.joseph@mila.quebec} \\ 
\small (Corresponding author)
}

\begin{document}
\maketitle

\begin{abstract}
While vision models are highly capable, their internal mechanisms remain poorly understood-- a challenge which sparse autoencoders (SAEs) have helped address in language, but which remains underexplored in vision. We address this gap by training SAEs on CLIP's vision transformer and uncover key differences between vision and language processing, including distinct sparsity patterns for SAEs trained across layers and token types. We then provide the first systematic analysis on the steerability of CLIP's vision transformer by introducing metrics to quantify how precisely SAE features can be steered to affect the model's output. We find that 10-15\% of neurons and features are steerable, with SAEs providing thousands more steerable features than the base model. Through targeted suppression of SAE features, we then demonstrate improved performance on three vision disentanglement tasks (CelebA, Waterbirds, and typographic attacks), finding optimal disentanglement in middle model layers, and achieving state-of-the-art performance on defense against typographic attacks.
\end{abstract}    
\section{Introduction}

Vision transformers have become fundamental to modern computer vision and have achieved remarkable performance across diverse tasks \citep{DBLP:journals/corr/abs-2010-11929, caron2021emerging, radford2021learning,liu2024visual}. However, despite their widespread adoption and success, we lack a deep understanding of how these models process and represent visual information internally. Although recent progress has been made in understanding the features of language models using decomposition techniques like sparse autoencoders (SAEs) \citep{cunningham2023sparseautoencodershighlyinterpretable, bricken2023towards}, similar interpretability advances in vision transformers remain limited.

\begin{figure}[!htb]
    \centering
    \includegraphics[width=0.47\textwidth]{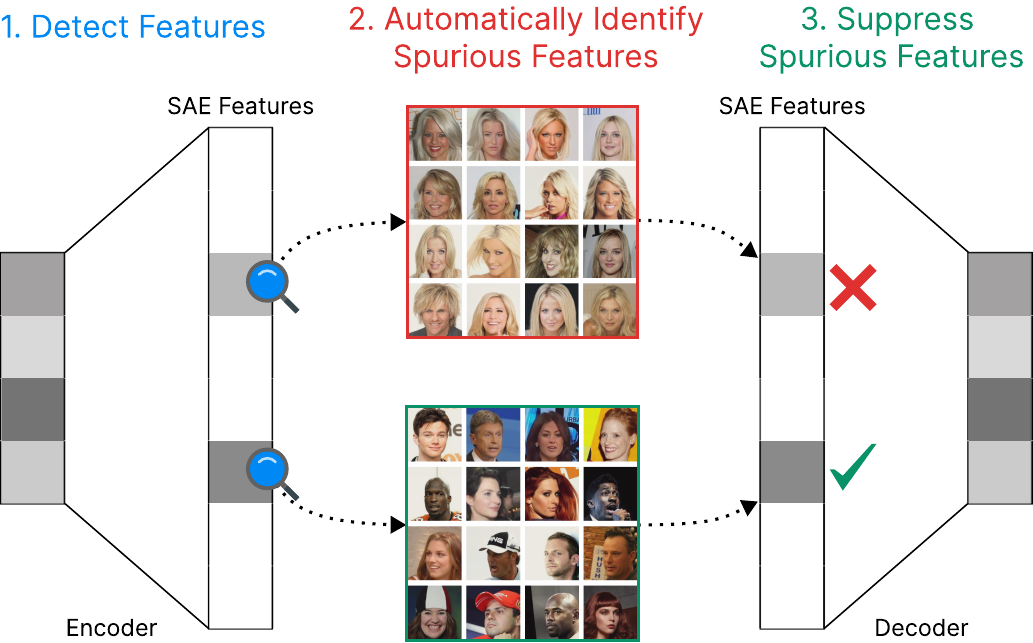}

    \caption{Our method improves performance on vision disentanglement tasks by  detecting and suppressing features with CLIP SAEs. For CelebA, we suppress blondeness to improve gender classification. For details, see Section \ref{ssec:supress_sc}.} 
    \label{fig:methods_figure}
\end{figure}

Our work addresses this gap by applying an SAE-based analysis to CLIP's vision transformer \citep{radford2021learning}, revealing fundamental differences between vision and language processing mechanisms. This analysis is particularly important as vision transformers increasingly serve as building blocks for larger multimodal systems \citep{liu2024visual,liu2024improved,dubey2024llama}, where understanding their internal representations becomes crucial for safety and reliability.

We make several key contributions. First, we train SAEs on CLIP's vision encoder. We observe properties about the sparsity of the resulting SAEs, as measured by the L0-norm, or the number of SAE features that activate for a given input token. The L0 values of SAEs trained on the spatial tokens are higher at the center of the image. Spatial tokens have much higher L0s than the CLS token and SAEs trained on a language model, suggesting fundamental differences in the sparsity of vision and language distributions. 


Continuing to characterize our SAEs, we introduce a steerability metric $S$, which measures how precisely SAE features can be manipulated to influence model outputs. Our analysis shows that approximately 10-15\% features are steerable (as determined by a threshold on $S$). While SAEs and the base model have similar proportions of steerable components, SAEs' higher dimensionality yields a much larger absolute number of steerable elements. The $S$ measurement provides a concrete framework for assessing the downstream utility of our SAEs.

Finally, we demonstrate the practical utility of our SAEs through improving the performance on three challenging vision disentanglement tasks: CelebA attribute separation, Waterbirds background suppression, and defense against typographic attacks. By selectively suppressing specific SAE features, we achieve improved performance on these tasks, with the most effective steering occurring in the middle layers of the model. For defense against typographic attacks, our method beats state-of-the-art performance.

Our findings not only advance our theoretical understanding of CLIP vision transformers but also provide practical tools for improving model behavior. We release our trained SAE models and code to facilitate further research in vision transformer interpretability. Our work establishes a foundation for more systematic approaches to understanding and controlling these increasingly important models.

\begin{figure}[h]
    \centering
    \includegraphics[width=\linewidth]{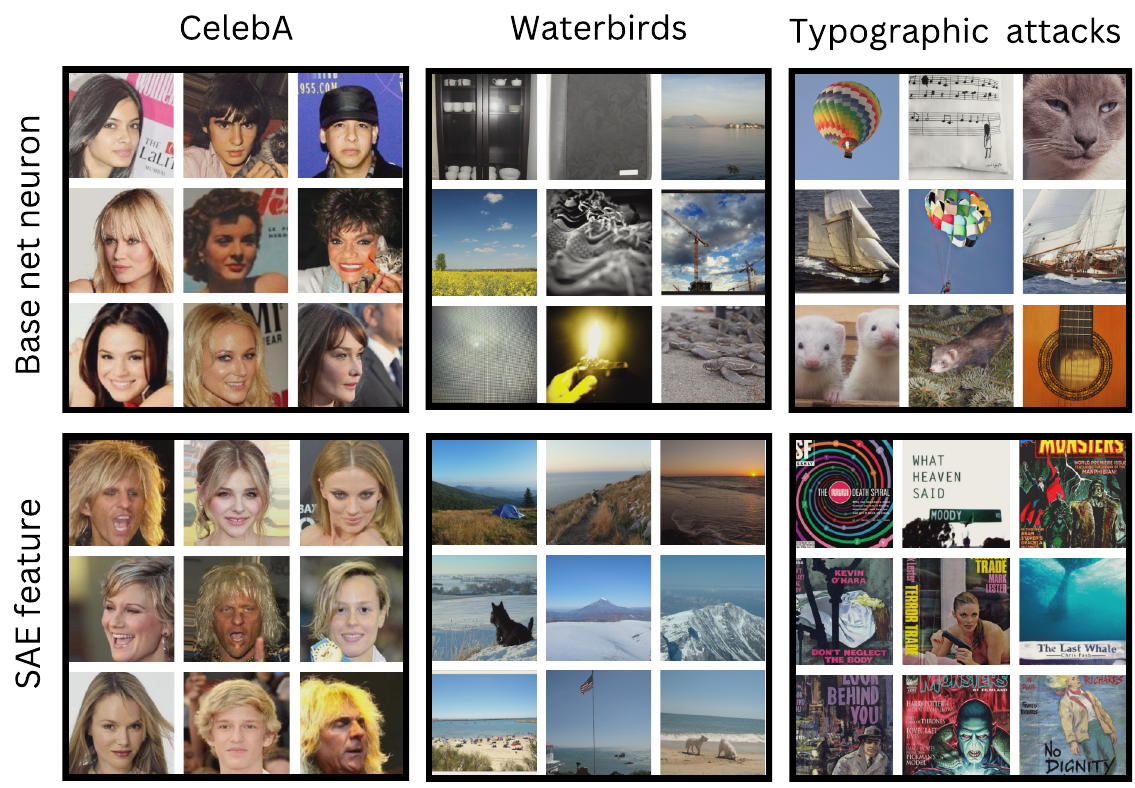}
    \caption{The top activating images show that base net features are polysemantic, while SAE features capture task-relevant attributes: blondeness (CelebA), land/water backgrounds (Waterbirds), and typographic images (typographic attacks). Feature selection details are in Section \ref{ssec:supress_sc} and more examples are in Appendix \ref{app:more_max_images}, Figure \ref{fig:FULL_SAE_neuron_max_act_images}.}
    \label{fig:SAE_neuron_max_act_images}
\end{figure}
\section{Background}

\subsection{CLIP-ViT preliminaries}

\subsubsection{Contrastive Pre-training}
CLIP \citep{radford2021learning} is trained via a contrastive loss to produce image representations from weak text supervision. The model includes an image encoder $M_{\text{image}}$ and a text encoder $M_{\text{text}}$ that map images and text descriptions to a shared latent space $\mathbb{R}^d$. The two encoders are trained jointly to maximize the cosine similarity between the output representations $M_{\text{image}}(I)$ and $M_{\text{text}}(t)$ for matching input text-image pairs $(t,I)$:
\begin{equation}
\text{sim}(I, t) = \frac{\langle M_{\text{image}}(I), M_{\text{text}}(t) \rangle}{\|M_{\text{image}}(I)\|_2 \|M_{\text{text}}(t)\|_2}.
\end{equation}
\subsubsection{Zero-shot Classification with CLIP}
Given a set of classes, the name of each class $c_i$ (e.g., the class ``tabby cat'') is mapped to a fixed $\text{template}(c_i)$ (e.g., ``A photo of a \{class\}''), and encoded via the text encoder $M_{\text{text}}(\text{template}(c_i))$. The classification prediction for a given image $I$ is the class $c_i$ whose text representation is most similar to the image representation:
\[
\arg\max_{c_i} \text{sim}(I,\text{template}(c_i)).
\]

\subsubsection{CLIP-ViT Architecture}
The CLIP-ViT image encoder consists of a Vision Transformer followed by a linear projection\footnote{The Vision Transformer (ViT) is applied to the input image $I \in \mathbb{R}^{H\times W \times 3}$ to obtain a $d'$-dimensional representation $\text{ViT}(I)$}. Denoting the projection matrix by $P \in \mathbb{R}^{d\times d'}$:
\begin{equation}
M_{\text{image}}(I) = P(\text{ViT}(I)).
\end{equation}
The input $I$ to ViT is first split into $K$ non-overlapping image patches that are encoded into $K$ $d'$-dimensional image tokens. An additional learned token, called the class (CLS) token, is included and used later as the output token. Tokens are processed simultaneously by applying $L$ alternating residual layers of multi-head self-attention (MSA) and MLP blocks.

\subsection{Feature disentanglement with sparse autoencoders}


One major challenge in neural network interpretability lies in the often non-interpretable and polysemantic nature of individual neurons \citep{elhage2022toymodelssuperposition}. Motivated by the sparse feature hypothesis \citet{olshausen1997sparse}, recent advancements in sparse dictionary learning \citet{cunningham2023sparseautoencodershighlyinterpretable,bricken2023towards} have shown that sparse autoencoders (SAEs) can effectively identify interpretable monosemantic directions in the latent space. For an input activation $x \in \mathbb{R}^{d_{\text{model}}}$ from these components, the SAE computes the following decomposition:
\begin{equation}
\hat{x} + \epsilon(x) = \sum_{j=1}^{d_{\text{SAE}}} f_j(x)n_j + b + \epsilon(x).
\end{equation}
This formulation decomposes the input into a reconstruction $\hat{x}$ through a sparse combination of feature vectors $n_j \in \mathbb{R}^{d_{\text{model}}}$, where each $n_j$ is normalized to unit length. The feature activations $f_j(x) \in \mathbb{R}$ serve as sparse coefficients, while $b \in \mathbb{R}^{d_{\text{model}}}$ represents a bias term. The model is optimized using a combination of $L_2$ reconstruction loss and $L_1$ regularization to enforce sparsity in the activations.

\section{Related Work}

\subsection{CLIP Interpretability}
Several works have explored interpreting CLIP's behavior and representations. Studies have investigated CLIP's biases \citep{singhal2022largelanguagemodelsencode}, attention layers \citep{gandelsmanclipdecomposition, joseph2024laying}, and neurons \citep{goh2021multimodal,gandelsman2024interpretingsecondordereffectsneurons}. Recent works \citep{bhalla2024interpreting,vielhaben2024beyond} introduce task-agnostic concept discovery, demonstrating that CLIP's internal representations naturally align with human-interpretable concepts. Our work builds upon these findings by using SAEs not only to understand CLIP's representations at a finer degree of resolution, but also to actively steer CLIP's behavior through identified interpretable features.

\subsection{Concept Bottleneck Models and Interpretability}
Recent efforts to make deep neural networks more interpretable have led to various approaches, with Concept Bottleneck Models (CBMs) emerging as a promising direction. Traditional CBMs \citep{koh2020concept} require manually specified concepts and labeled attribute datasets, limiting their scalability. Recent work leverages large language models (LLMs) and vision-language models to overcome this limitation \citep{oikarinen2023labelfree,panousis2023sparselinearconceptdiscovery,chattopadhyay2023information}. However, these approaches are subject to the biases of LLMs and still rely on pre-selecting concepts based on downstream tasks, which may not align with the model's learned representations. In contrast, our work uses SAEs, which extract unsupervised features from the model's internal representations.

\subsection{Sparse Autoencoders for Model Interpretability}
Sparse autoencoders (SAEs) have recently gained attention as tools for mechanistic interpretability \citep{cunningham2023sparseautoencodershighlyinterpretable,bricken2023towards}. These works demonstrate that SAEs can effectively decompose neural networks into interpretable features, particularly in language models.

Recent studies have expanded the application of SAEs to vision tasks. \citet{fry2024multimodal,abdulaal2024x} and \citet{daujotas2024interpreting, daujotas2024case} explored how SAEs can extract interpretable features from CLIP’s vision encoder, and Daujotas showed their potential in modifying image generation with diffusion models. Meanwhile, \citet{rao2024discoverthennametaskagnosticconceptbottlenecks} leveraged CLIP embeddings to label SAE-derived concepts, enabling task-agnostic concept bottleneck models. \citet{gorton2024the} applied SAEs to InceptionV1, uncovering missing curve detectors. Additionally, \citet{lim2024sparseautoencodersrevealselective} introduced PatchSAE for spatially localized attributions for fine-grained concept extraction.   

Our paper builds upon past insights, uncovering novel vision-specific sparsity patterns in the SAEs' feature space, performing the first systematic and quantitative analysis on the steerability of CLIP's features, and applying CLIP SAEs to practical disentanglement tasks, achieving state-of-the-art performance on defense against typographic attacks.

\begin{figure}
    \centering
    \includegraphics[width=0.8\linewidth]{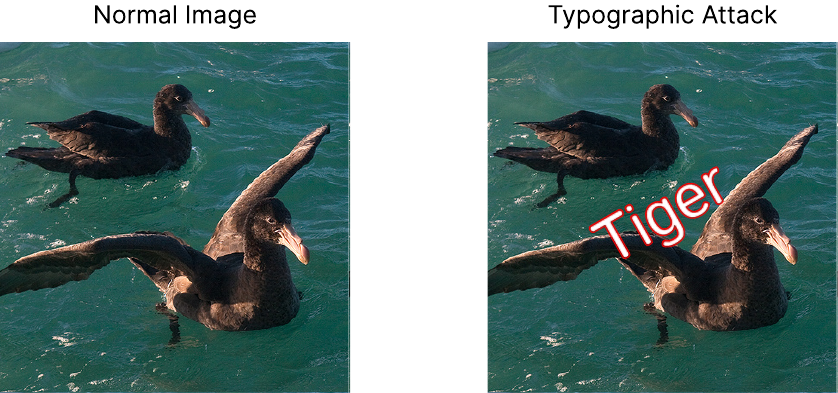}
    \caption{Typographic Attack on CLIP: On the left, an ImageNet-100 sample. On the right, the same image with 'tiger' written on it. As demonstrated by \citet{goh2021multimodal}, this simple text overlay can mislead CLIP's zero-shot classification towards the attacker's intended label.}
    \label{fig:typo-attack}
\end{figure}

\subsection{Typographic Attacks in CLIP Models}
Typographic attacks \cite{goh2021multimodal} are a specific attack vector targeting vision-language models like CLIP, arising from their multimodal pretraining. These attacks involve inserting human-readable text into an input image to manipulate the model’s prediction, as illustrated in Figure \ref{fig:typo-attack}.

To improve robustness against such attacks, \citet{materzynska2022disentangling} proposed a learned transformation applied to CLIP's output. PAINT \cite{ilharco2022patching} fine-tunes on synthetically generated images of typographic attacks and uses weight interpolation of the fine-tuned and the original model, to gain a more robust model. \citet{azuma2023defense} introduced Defense-Prefixes (DPs) which extend the idea of ``class-prefix learning'' \cite{ruiz2023dreambooth,kumari2023multi,lin2023magic3d,schwartz2023discriminative} towards defending against typographic attacks by inserting a learned token before the class name, leading to improved robustness without changing the model's parameters.

\section{Training CLIP Vision SAEs}

We train two variants of SAEs on activations from the residual stream of each layer of the CLIP-ViT-B-32 model. The Vanilla SAEs use the ReLU activation function with the sparsity induced by L1 regularization, while the Top-K SAEs \cite{gao2024scalingevaluatingsparseautoencoders} use the Top-K activation function with a fixed sparsity as defined by the hyperparameter \textit{k}. Both sets of SAEs are trained in general and task-specific settings. The general purpose SAEs are trained on ImageNet-1K to populate the dictionary with as many features as possible from the ViT's latent space, whereas the task-specific SAEs are trained on CelebA and Waterbirds datasets to populate the dictionary with features that are task-relevant. 
Full training and evaluation details of our SAEs are in Appendix \ref{app:sae_training_details}.

\section{Properties of CLIP SAEs} 

\begin{figure*}[!htb]
    \centering
    \includegraphics[width=\textwidth]{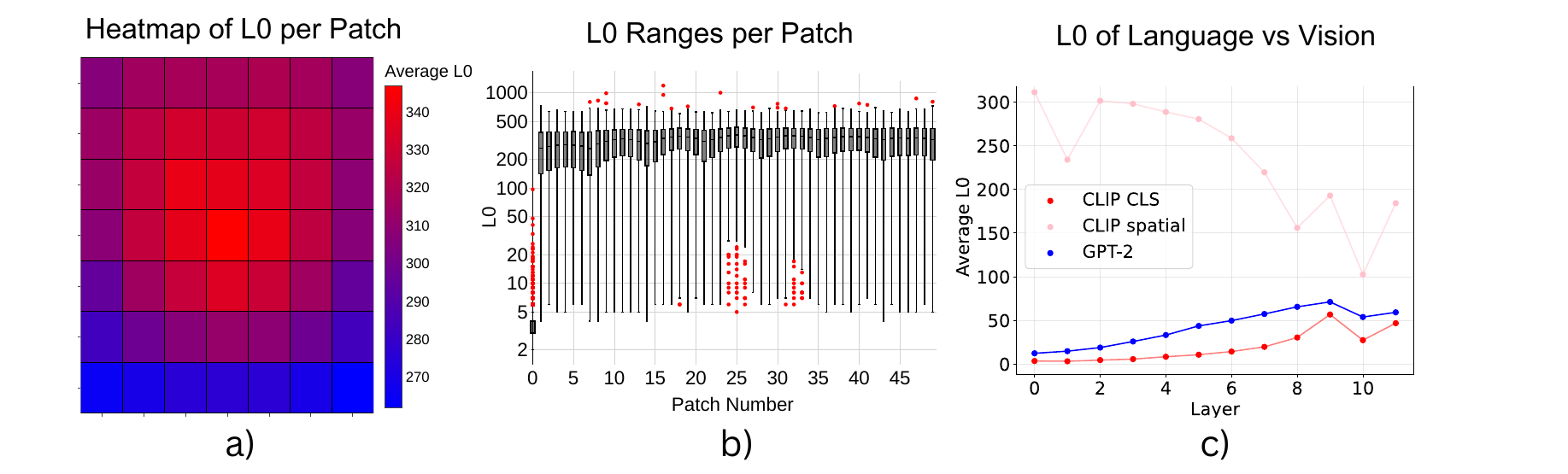}
    \caption{A visualization of the L0 values for an x64 vanilla SAE trained on all patches of CLIP-B-32 for Layer 0. a) A heatmap of average L0 per patch, overlaid on the original image grid, shows that there is a bias toward the center. The center bias remains constant for all layers (see Appendix \ref{app:l0_comparison_details}). b) A box plot of L0s per patch reflects high-norm spatial tokens and a low-norm CLS token. c) A comparison between the L0s SAEs trained on the residual stream of GPT-2 and CLIP.} 
    \label{fig:l0_analysis}
\end{figure*}

\subsection{Comparing Spatial and CLS Token Behaviors}

\begin{figure}[!h]
\centering
\caption{Heatmap of average L0 per layer shows patch sparsity still retains its spatial bias even in Layer 11.}
\includegraphics[width=0.25\textwidth]{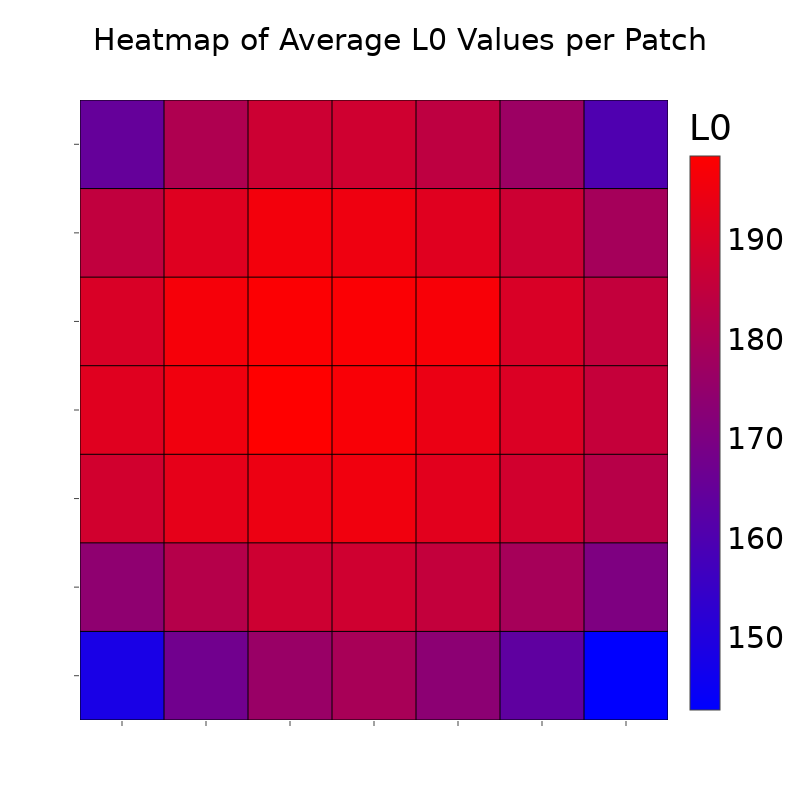}
\label{l0_layer_11}
\end{figure}

The L0 metric, which measures the number of activated features per patch and serves as a key indicator of SAE sparsity, reveals distinct activation patterns in CLIP SAEs. Spatial tokens maintain high activation counts (300-700), with central patches showing a higher L0, suggesting greater information density in these regions (Figure \ref{fig:l0_analysis} a). The spatial bias in L0 persists up through the final layer, showing that some spatial information is retained (Appendix \ref{app:l0_comparison_details}, Figure \ref{l0_layer_11}). The CLS token follows a distinct pattern from the spatial tokens, transitioning from sparse to rich representations around layers 6-7 (Figure \ref{fig:l0_analysis} b, c). The L0 of the spatial tokens has a high degree of variance, reflecting high norm tokens (Figure \ref{fig:l0_analysis}).



\subsection{Comparing CLIP and Language Model Sparsity}
When contrasting CLIP with GPT-2, we find striking differences in the L0 values of the SAEs trained on their representations (Figure \ref{fig:l0_analysis}c). CLIP's spatial tokens maintain substantially higher L0 values, showing 3-14x higher activation counts than GPT-2 tokens (which average L0 = 20-50). Interestingly, CLIP's CLS token exhibits sparsity patterns more similar to those of language models in early layers before diverging.

These patterns suggest fundamental differences in information processing strategies. While language models maintain relatively consistent sparsity throughout their layers, CLIP demonstrates a dual behavior: spatial tokens preserve rich local features while the CLS token captures the compression, which aligns with observations in \cite{achtibat2024attnlrp}. 


\section{CLIP SAE Feature Steering}

Building on our analysis of CLIP SAEs, we explore their potential for model control through feature activation steering—manipulating SAE features to influence model outputs predictably \citep{templeton2024scaling}. This investigation raises several key questions: how to measure the impact of feature manipulations on outputs, how SAE steering compares to direct model steering, how steering effectiveness varies across layers, and what proportion of features can be effectively steered.

To address these questions, we introduce a metric called \textit{steerability}, which quantitatively characterizes an SAE feature's steering performance. This metric measures output probability distribution changes in response to changing the activation of a given feature, enabling systematic comparisons between SAE features, across SAE layers, and between SAEs and their base networks. Below, we define steerability metrics and use them to analyze properties of our CLIP SAEs.

\subsection{Steerability Metrics} \label{six_one}
We conduct our analysis in the context of zero-shot CLIP classification. Following \citep{rao2024discoverthennametaskagnosticconceptbottlenecks}, we use CLIP's text encoder to encode a vocabulary $\mathcal{V}={v_1, v_2, \ldots, v_m}$ containing thousands of potential feature names into a library of text embeddings $\mathcal{T}={t_1, t_2, \ldots t_m}$.


For a given image $i \in I$, we first obtain its embedding $i_\text{emb}$ through CLIP's vision encoder. Computing the dot product between $i_\text{emb}$ and $\mathcal{T}$ followed by a softmax operation yields CLIP's probability distribution $P_i \in \mathbb{R}^{|\mathcal{V}|}$ over the vocabulary $\mathcal{V}$.
To measure steering effects, we then select a feature $f$ and replace its feature activation across all patches with a steering strength $s$ during the forward pass. Let $\Tilde{P}_i \in \mathbb{R}^{|\mathcal{V}|}$ CLIP's steered probability distribution over the vocabulary $\mathcal{V}$.

We now define feature-level metrics that help us to compare the steering capabilities of different features, and furthermore propose metrics to help us to compare SAE feature-level and neuron-level representations with regard to steerability.

\subsubsection{Feature Level Steerability Metrics}


The average probability difference of a feature $f$ across images $\mathcal{I}$ is given by:
\begin{equation}
\Delta P_f = \left|\frac{1}{|\mathcal{I}|} \sum_{i \in \mathcal{I}} (\tilde{P}_i - P_i) \right|, \label{eq:delta_p}
\end{equation}
where $\tilde{P}_i$ and $P_i$ are the predicted probabilities after and before feature activation, respectively.

Furthermore, we quantify $f$’s steerability by:
\begin{equation}
S_f = \left|\frac{1}{|\mathcal{I}|} \sum_{i \in \mathcal{I}} (\tilde{P}_i - P_i)^2 \right|. \label{eq:sae_steer}
\end{equation}

$\mathcal{S}_f$ is designed to quantify how a feature distributes its weight across concept vocabulary elements. For a feature with probability mass uniformly distributed among its top $n_c$ concepts with value $\frac{1}{n_c}$, the metric has the property: $\sum_{i=1}^{n_c}\left(\frac{1}{n_c}\right)^2 = n_c \cdot \left(\frac{1}{n_c}\right)^2 = \frac{1}{n_c}$. This formulation ensures that polysemantic features with approximately uniform distributions will have the desirable property that a feature with $n_c - 1$ top concepts always scores higher than one with $n_c$ top concepts.  

\subsubsection{Layer Level Steerability Metrics}
The layer-level metrics provide a more comprehensive view of steerability across the features within a layer. First, we define the average steerability of a layer, which reflects the overall steerability of all features in that layer: \begin{equation} \mathcal{S} = \dfrac{1}{|F|} \sum_{f \in F} S_f, \label{eq:layer_steer} \end{equation} where $F$ denotes the set of all features in the layer and $S_f$ is the steerability of feature $f$. This metric aggregates the individual steerabilities of all features to give a sense of how well the layer as a whole can be steered.



To identify how many features within a layer are steerable, we define the number of steerable features as the proportion of features whose steerability exceeds a given threshold $\gamma$: \begin{equation} \#\mathcal{S} = \dfrac{1}{|F|} \sum_{f \in F} \mathbf{1}[S_f > \gamma], \label{num_feats_sae} \end{equation} where $\mathbf{1}[\cdot]$ is the indicator function. This metric indicates how many features exhibit significant steerability, characterizing how steering a given layer impacts the model’s output.

Finally, we quantify concept coverage as the number of features whose steerability exceeds a higher threshold $\beta$, corresponding to a meaningful increase in concept probability: \begin{equation} \#\mathcal{C} = \sum_{f \in F} \mathbf{1}[S_f > \beta]. \label{n_a_c} \end{equation}

The $\#C$ metric only indicates the number of vocabulary concepts a feature adheres to, without capturing the semantic relationships between promoted concepts. Thus, 
a feature may exhibit high specificity while remaining polysemantic (e.g., equally promoting 
semantically distant concepts like ``carrot'' and ``organization''). We elaborate on this limitation in Appendix \ref{app:limitations_steerability} and leave further exploration to future work.

\subsection{Empirical Results on CLIP SAE Steerability}

Our analysis reveals three key findings about steerable features in CLIP shown in Figures \ref{fig:afsp}, \ref{fig:histogram_steerability}, and \ref{feat_vs_neurons}. First, we discover features that can strongly influence CLIP's output distribution when steered, including ``perfectly'' steerable features that direct all probability mass towards one concept when steered to its asymptote. Second, using our layer-level metrics, we quantify how common these steerable features are. In CLIP's deeper layers, typically 10-15\% of features are steerable, where $\#\mathcal{S} > |F| \cdot 0.10$ (Eq. \ref{num_feats_sae}). Finally, we empirically show that steering at the feature-level provides much better control over concept space than neuron-level steering, achieving more than 10 times the concept coverage $\#\mathcal{C}$ (Eq. \ref{n_a_c}) compared to the base model.


\subsubsection{Some Properties of Steerable Features}
\begin{figure}[t!]
\includegraphics[width=\linewidth]{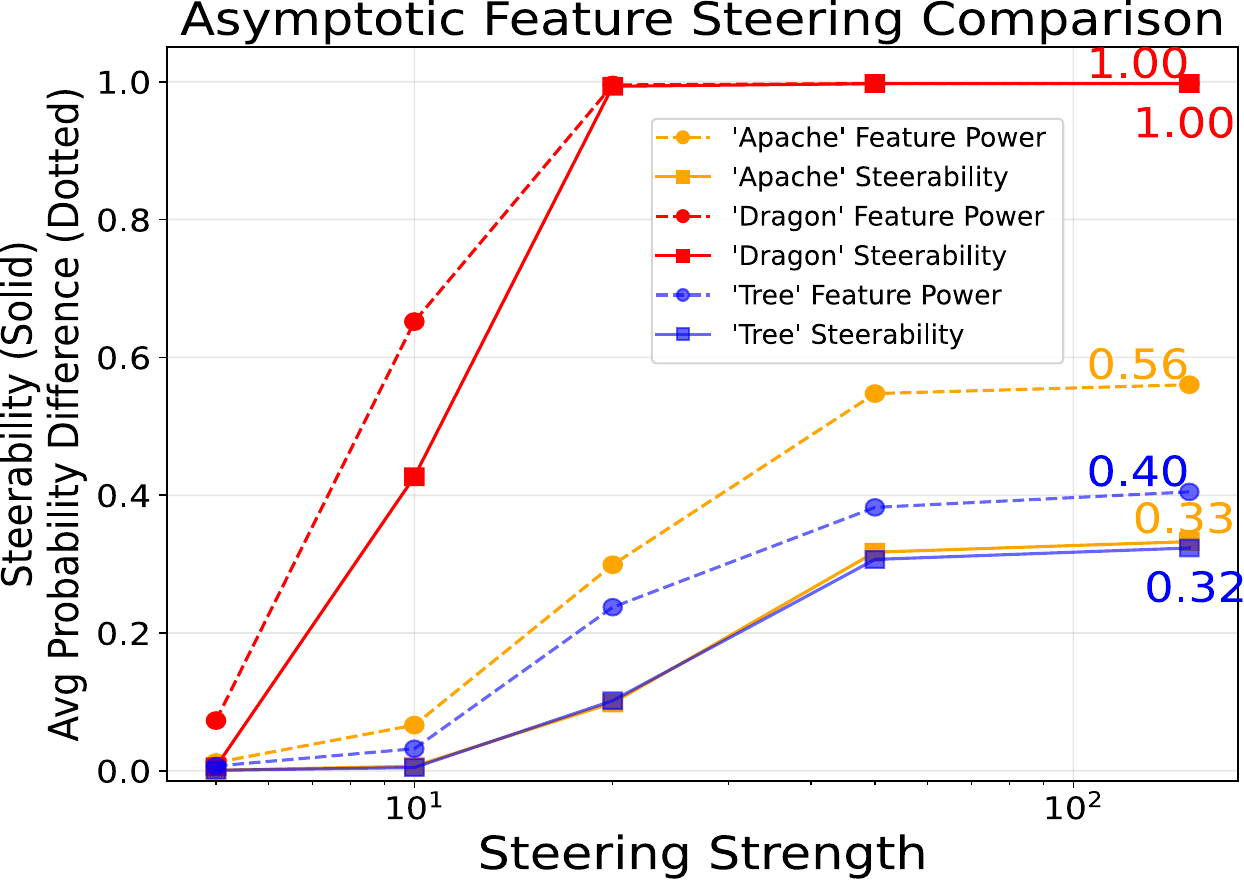}
\caption{Asymptotic Feature Steerability Plot showing $\Delta P_f$ (dotted) and $\mathcal{S}_f$ (solid) versus steering strength. The ``dragon'' feature achieves perfect steering to a single concept, while ``tree'' and ``apache'' have similar $\mathcal{S}_f$ but different $\Delta P_f$ - 'tree' steers precisely to tree concepts, whereas ``apache'' disperses across helicopter-related concepts (e.g., ``aircraft'', ``aviation'', ``rescue'').}

\label{fig:afsp}
\end{figure}

When we amplify features to their maximum strength\footnote{We determine a value of 150 to be sufficient, as larger values produced no additional changes. This threshold may vary by SAE and base model.}, their effect on probability distribution varies. Some features, like the ‘Dragon’ feature shown in Figure \ref{fig:afsp}, strongly concentrate probability mass toward a single concept—in this case, the dragon logit—or a small cluster of related concepts. Other features, however, have a more diffuse effect: some appear to have little to no impact on the probability distribution, while others spread probability mass across a broad range of concepts, sometimes affecting hundreds or even thousands of them.

\begin{figure}[H]
\caption{Feature Concept Space Throughout Steering. This is a plot of the top three logits for each steering strength run. This visualizes the causal link between steering strength and CLIP prediction, and indicates whether a feature is semantically coherent or polysemantic.}
\centering
\label{fig:feature_concept_space}
\includegraphics[scale=0.3]{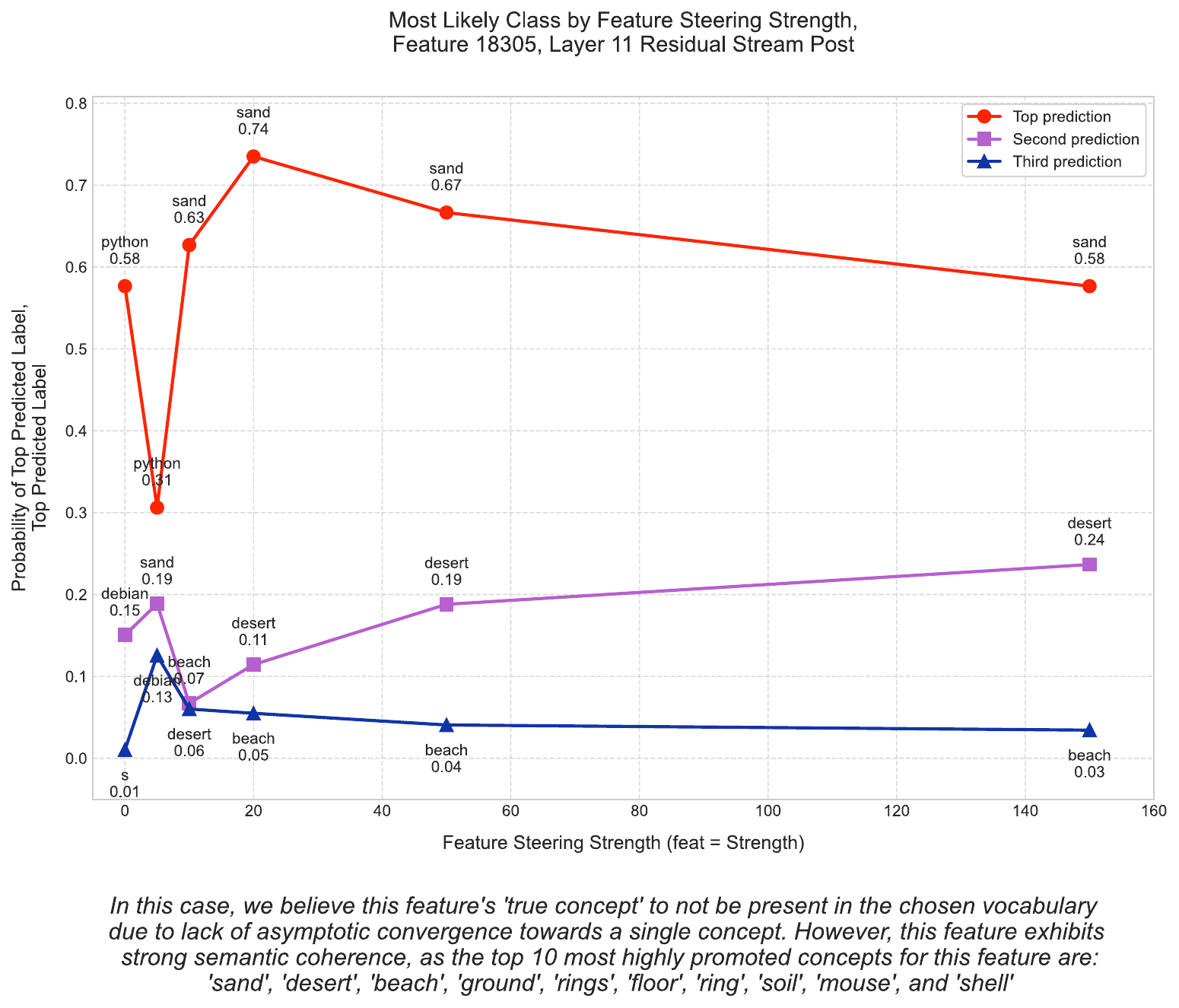}
\end{figure}

This second pattern could have several explanations: these features might not meaningfully affect CLIP's output, they might need to work in concert with other features (suggesting our steering pushes activations too out-of-distribution), or our vocabulary might lack terms that accurately capture the feature's true semantic direction in CLIP's concept space (Appendix \ref{app:steering}, Figure \ref{fig:feature_concept_space}) \footnote{We choose $\#\mathcal{V} = 5,000$ to balance coverage and compute; a larger vocabulary could mitigate this limitation.}.


\subsubsection{The Frequency of Steerable Features}
\begin{figure}[h]
\centering
\includegraphics[scale=0.3]{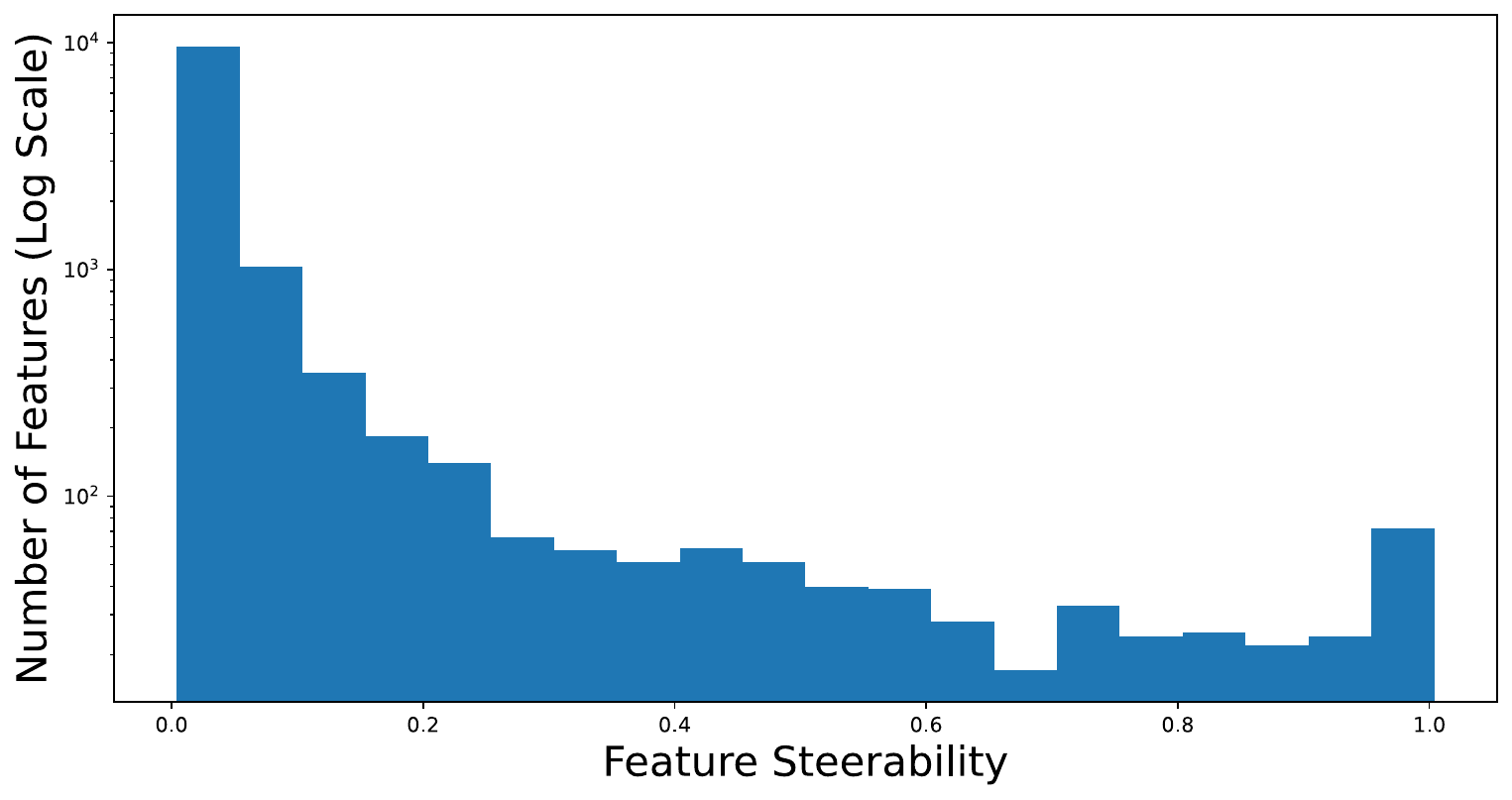}
\caption{Log-Scale SAE Feature Steerability Histogram. Steerability scores at $s$ = 150.0 for a Vanilla SAE trained on the residual stream of Layer 11. This is a sample of 12,000 features (representing roughly 25\% of a total 49,152), of which 1,322 are steerable ($\mathcal{S}_f$ $>$ 0.10).}
\label{fig:histogram_steerability}
\end{figure}

Using Eq. \ref{eq:sae_steer}, we analyze the prevalence of steerable features in an SAE. Figure \ref{fig:histogram_steerability} presents our findings for a Vanilla SAE trained on layer 11 of CLIP’s residual stream. Due to computational constraints, we analyze a subset of 12,000 features. 
 Using a steerability threshold of $\gamma = 0.10$ we find 1,322 steerable features among 12,000 total features. While these constitute a minority, they still form a substantial subset of controllable features, including approximately 100 that are ``perfectly'' steerable. This metric provides a valuable tool for practitioners to optimize SAE training, increasing the number of steerable features and identifying suboptimal training outcomes.


\subsubsection{Feature Steering vs. Neuron Steering}
\begin{figure}[h]
\centering
\includegraphics[scale=0.28]{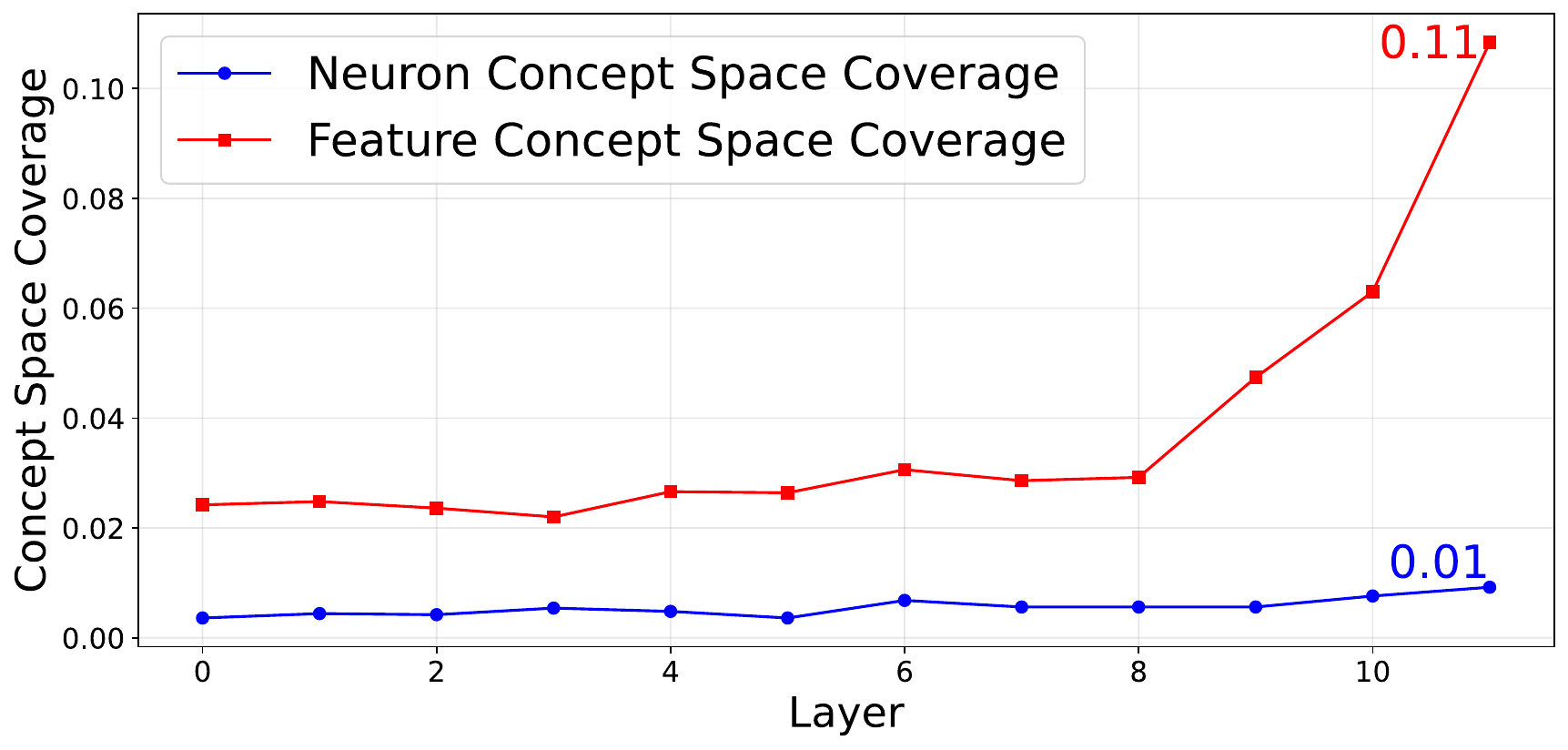}
\caption{Concept Space Coverage, Feature vs Neurons, by Layer (Residual Stream). Notably, at layer 11, feature steering allows access to more than 10x the span of concept space that neuron steering does.}
\label{feat_vs_neurons}
\end{figure}

Our comparison of feature-level and neuron-level steering reveals a significant advantage for feature-level. While individual steerable neurons can match the steering strength of features, they are far rarer and provide access to a much smaller concept space (Figure \ref{feat_vs_neurons}).
In a sample of 25\%  of features and neurons from layer 11, neuron steering accessed only 1\% of concepts (about 50 concepts from our 5000-word vocabulary) in $\mathcal{V}$ (Eq. \ref{n_a_c}), while feature steering accessed approximately 11\% (about 550 concepts).

\begin{table*}[ht]
\centering
\small
\begin{tabular}{l|cc|cc|cc|cc}
\toprule
& \multicolumn{4}{c|}{\textbf{CelebA}} & \multicolumn{4}{c}{\textbf{Waterbirds}} \\
\cmidrule(lr){2-5} \cmidrule(lr){6-9}
& \multicolumn{2}{c|}{Original Model} & \multicolumn{2}{c|}{SAE} & \multicolumn{2}{c|}{Original Model} & \multicolumn{2}{c}{SAE} \\
\cmidrule(lr){2-3} \cmidrule(lr){4-5} \cmidrule(lr){6-7} \cmidrule(lr){8-9}
& Overall & Worst & Overall & Worst & Overall & Worst & Overall & Worst \\
\midrule
Baseline & 92.78 & 77.78 & - & - & 68.81 & 22.43 & - & - \\
\midrule
Layer 0 & 92.78 & 78.89 & \textbf{93.19} & \textbf{79.44} & 68.81 & 22.43 & \textbf{69.42} & 22.43 \\
Layer 1 & 92.78 & 77.78 & \textbf{92.92} & \textbf{78.89} & 68.81 & 22.43 & 68.81 & 22.43 \\
Layer 2 & 92.78 & 77.78 & \textbf{93.47} & \textbf{79.44} & 68.81 & 22.43 & \textbf{71.95} & *\textbf{24.61} \\
Layer 3 & 92.78 & 77.78 & \textbf{93.06} & \textbf{78.89} & 68.81 & 22.43 & \textbf{69.54} & \textbf{24.14} \\
Layer 4 & 92.78 & 77.78 & \textbf{93.19} & \textbf{80.00} & 68.81 & 22.43 & \textbf{68.93} & \textbf{23.21} \\
Layer 5 & 92.78 & 77.78 & \textbf{93.19} & \textbf{78.89} & 68.81 & 22.43 & \textbf{69.05} & \textbf{22.74} \\
Layer 6 & 92.78 & 77.78 & \textbf{93.19} & \textbf{80.00} & 68.81 & 22.43 & 68.81 & 22.43 \\
Layer 7 & 92.64 & 77.78 & \textbf{93.61} & *\textbf{81.11} & 68.81 & 22.43 & 68.81 & 22.43 \\
Layer 8 & 92.64 & 77.78 & \textbf{93.47} & *\textbf{81.11} & 68.81 & 22.43 & 68.81 & 22.43 \\
Layer 9 & 92.78 & 77.78 & \textbf{93.06} & \textbf{79.44} & 68.81 & 22.43 & \textbf{70.19} & \textbf{22.74} \\
Layer 10 & 92.78 & 77.78 & \textbf{93.47} & \textbf{80.00} & 68.81 & 22.43 & 68.81 & 22.43 \\
Layer 11 & 92.50 & 77.22 & \textbf{93.33} & \textbf{80.56} & 68.81 & 22.43 & 68.81 & 22.43 \\
\bottomrule
\end{tabular}
\caption{Performance comparison showing overall and worst-group accuracy across different layers and tasks after targeted zero-ablations on the original model neurons (control), and targeted SAEs zero-ablations of the strict feature set $F^l_{\tau^*}$ (our method) (See Section \ref{ssec:supress_sc}). Ablations using SAE feature outperform ablations on the base model, due to SAE's finer level of granularity in representing the spurious feature. We pick the top-performing SAE for each layer and task. Full results for all SAE types, random controls, and the relaxed condition are in Appendix \ref{app:disentanglement_tasks}.}
\label{tab:layer_analysis}
\end{table*}

\section{Improving on Disentanglement Tasks with CLIP SAE Feature Steering}



In this section, we apply our CLIP SAEs to three distinct downstream tasks designed to assess their disentanglement capabilities and demonstrate the universality of the learned features.

Our first objective is to evaluate the effectiveness of SAEs in suppressing spurious correlations, drawing inspiration from frameworks such as \citet{pahde2022navigating} and \citet{dreyer2023hope}. To achieve this, we adopt experimental settings similar to those used in \citet{nam2022spread}, \citet{idrissi2022simple}, and \citet{pham2021effect}:
\begin{enumerate}
    \item \textbf{Waterbirds}~\cite{sagawa2019distributionally}: is an artificially generated dataset which combines bird photographs in the Caltech-UCSD Birds dataset\cite{wah2011caltech} with background images from the Places dataset\cite{zhou2017places}. The goal is to classify the binary target attribute $\quad Y = \text{waterbird} \quad \text{and} \quad \overline{Y} = \text{landbird}\quad$ given the spurious correlations with the background landscape $\quad A = \text{water background} \quad \text{and} \quad \overline{A} = \text{land background}\quad $ 
    \item \textbf{CelebA}~\cite{liu2018large}: consists of the face pictures of celebrities, featuring annotations on a variety of features. We use the binary gender label\footnote{While we acknowledge that gender is not binary, we follow this benchmark as it is standard in the literature.} as the target attribute $\quad Y = \text{male} \quad \text{and} \quad \overline{Y} = \text{female}\quad$ and utilize the feature $\quad A = \text{blond} \quad \text{and} \quad \overline{A} = \text{not blond}\quad$ as a spurious correlation.
\end{enumerate}

Secondly, we aim to assess the utility of our SAEs in safety-critical applications by evaluating their effectiveness in enhancing the robustness of the CLIP model against typographic attacks. To this end, we follow the evaluation setup of \citet{azuma2023defense}, using ImageNet-100\footnote{\url{https://www.kaggle.com/datasets/ambityga/imagenet100}}, a subset of the ImageNet dataset \cite{deng2009imagenet}, for tuning our method. We then benchmark our approach on RTA-100 \cite{azuma2023defense}, PAINT \cite{ilharco2022patching}, and the dataset published by \citet{materzynska2022disentangling}.


\subsection{Suppressing Spurious Correlations}\label{ssec:supress_sc}
To suppress the spurious correlations, we identify the SAE features that are most closely aligned to the spurious correlation $A$ (Figure \ref{fig:methods_figure}). To achieve this, we split the train dataset $D$ into two subsets, $D_A$ and $D_{\overline{A}}$, based on whether the spurious correlation is present in the datapoint. Then, for each layer $l$, select the SAE features $F^l$ whose average activation on $D_A$ is at least $\tau$ higher than their average activation on $D_{\overline{A}}$. Formally,
\begin{equation}\label{eq:feature_set}
F^l = \bigl\{ j \: \big| \; \mathbb{E}_{x \sim D_A} [f_j^l(x)] \;>\; \mathbb{E}_{x \sim D_{\overline{A}}}[f_j^l(x)] \; + \; \tau\ \bigr\},
\end{equation}
where $f_j^l(\cdot)$ is the activation of the $j$th SAE feature of the $l$th layer, and $\tau \in \mathbb{R}$ is a scalar threshold. 

We evaluate the CLIP model on the validation set, applying zero ablation to the set of SAE features $F^l$ independently for each layer.

To systematically determine an appropriate threshold, we perform a grid search over $\tau \in [10^{-6}, 1.0]$ for each layer $l$. For each threshold value, we obtain a corresponding feature set, denoted as $F^l_\tau$. These feature sets are evaluated to assess their impact using two metrics: $F^l_{\tau^*}$, which maximizes both overall accuracy and worst-group accuracy; and $F^l_{\tau'}$, which allows for a controlled performance drop ($\leq 4\%$ drop in accuracy) in non-target groups while prioritizing improvements in the worst-performing group.

These thresholds $F^l_{\tau^*}$ and $F^l_{\tau'}$ are chosen to balance overall model performance with fairness considerations. The 4\% threshold is chosen to ensure that gains for the worst-performing group are achieved without excessive degradation in other groups.


\subsubsection{Evaluation}
We evaluate the CLIP model on the held out test set, while ablating the SAE features sets $F^l_{\tau^*}$ and $F^l_{\tau'}$. Additionally, we apply the same technique to CLIP's original feature space to verify that the SAE enhances disentanglement compared to the unmodified CLIP representation. To further validate our approach, we perform random ablations in both the SAE and original CLIP feature spaces, ensuring that our method accurately identifies the relevant features.

\subsubsection{Results}

Targeted feature suppression with SAEs consistently improves disentanglement for every layer of the model for CelebA, and for many layers for Waterbirds (Table \ref{tab:layer_analysis}). Based on maximally activating images for $F^l_{\tau^*}$, the base model neurons are highly polysemantic, while SAE features show the task-relevant attribute (Figure \ref{fig:SAE_neuron_max_act_images}).

The optimal layer for disentanglement is Layers 7 and 8 of CelebA, with worst group accuracy improving from 77.78\% to 81.11\%, and Layer 2 for Waterbirds, with worst group accuracy improving from 22.43\% to 24.61\% (Table \ref{tab:layer_analysis}). Under the relaxed condition, the best-performing accuracy goes up to 86.67\% at Layer 9 for CelebA (Appendix \ref{app:disentanglement_tasks}, Table \ref{table:celeba_total_results_relaxed}) and to 36.6\% at Layer 5 for Waterbirds (Appendix \ref{app:disentanglement_tasks}, Table \ref{table:waterbirds_total_results_relaxed}). We hypothesize that more diffuse features (like land and water backgrounds) may be optimally disentangled in earlier layers, before they become entangled with other features during the forward pass. More localized features (like blonde hair) may be better disentangled in later layers.





\subsection{Suppressing Typographic Attacks}
\label{sec:typographic_attack}

\begin{table}[htbp]
\centering
\small
\begin{tabular}{lccc}
\toprule
Defense  & RTA 100 & PAINT & \shortstack{\citet{materzynska2022disentangling}} \\
\midrule
No Defense       & 0.66  & 0.63 & 0.54      \\
DP               & \textbf{0.73} & 0.66 & 0.83    \\
SAE (Ours)       & 0.72 & \textbf{0.67} & 0.79      \\
\midrule
DP + Ours        & \textbf{0.73} & \textbf{0.67} & \textbf{0.88}    \\
\bottomrule
\end{tabular}
\caption{Performance comparison of our method, Defense-Prefix (SoA), and their combination across three standard benchmarks for typographic attacks. See more in Section \ref{sec:typographic_attack}}.
\label{tab:defense_comparison}
\end{table}

To obtain a set of SAE features that encode typographic knowledge, we follow a similar approach to Section~\ref{ssec:supress_sc}. For clarity, we focus our study of typographic attacks on CLIP’s last residual layer. Throughout the remainder of this section, we omit the layer superscript (e.g., $^l$) for readability.

We define two datasets: ImageNet-100 serves as $D_{\overline{T}}$, while $D_T$ is constructed by applying a synthetically generated typographic attack to each image in ImageNet-100. 
Accordingly, the SAE feature set $F$ is obtained following Equation~\ref{eq:feature_set}.

Next, we construct a simple extension to $F$ to improve the recall of typographic features. Specifically, for each feature direction $n_j$, we check whether it has a cosine similarity higher than $\lambda$ with any $n_m$ for $m \in M = \{ m \mid f_m \in F \}$. If so, it is added to an enhanced set of features $F^*$, thereby capturing additional relevant features that may not have been initially selected.

\subsubsection{Evaluation}

Similar to Section~\ref{ssec:supress_sc}, we perform a sweep over the $\lambda$ and $\tau$ values on the validation set. The test results on the three benchmarks—RTA-100 \cite{azuma2023defense}, PAINT \cite{ilharco2022patching}, and the dataset published by \citet{materzynska2022disentangling}—are reported in Table \ref{tab:defense_comparison}.

Additionally, we report the results of \citet{azuma2023defense}, who train a Defense-Prefix (DP) to improve CLIP’s robustness against typographic attacks, as it represents the current state-of-the-art (SOTA). We also evaluate the combination of our method with DP. This integration is straightforward, as DP modifies only the input to CLIP’s text transformer, while our approach exclusively affects the vision transformer.

\subsubsection{Results}We conduct our tests with $\lambda = 0.2$ and $\tau = 1$, which result in $F^*$ containing 495 out of 49,152 possible SAE features ($\sim1\%$). The drop in ImageNet top-1 accuracy was 0.7 percentage points. Our method outperforms the current state-of-the-art on the PAINT dataset and remains competitive with DP on the \citep{azuma2023defense} and \citet{materzynska2022disentangling} datasets. When combining our method with the SOTA, we achieve top performance across all benchmarks (Table \ref{tab:defense_comparison}).

\section{Conclusion}
In this work, we train sparse autoencoders (SAEs) on CLIP's vision transformer, revealing sparsity differences between vision and language processing. We find that 10-15\% of features in deeper layers are steerable, with CLIP SAE features providing 10x better concept coverage than base neurons. Through these steerable features, we demonstrate improved disentanglement capabilities and achieved state-of-the-art performance on typographic attack defense. 


{
    \small
    \bibliographystyle{ieeenat_fullname}
    \bibliography{main}
}
\setcounter{page}{1}

\maketitlesupplementary

\section{L0 comparison details}
\label{app:l0_comparison_details}

The L0 values for language SAEs were collected from open source sparse autoencoders trained on GPT2-s residual stream \cite{josephbloomsae}.

\section{SAE training details and statistics}
\label{app:sae_training_details}

\subsection{Training procedure}

\textbf{Architecture and Optimization.}
The SAEs were configured with an expansion factor of 64, mapping the base model’s 768-dimensional activation space to a dictionary of 49{,}152 features. For initialization, we set the encoder weights to be the transpose of the decoder weights. The SAEs were trained using the Adam optimizer, sweeping the initial learning rate from 1e-5 to 1e-1, and employed a cosine annealing learning rate schedule with a 200-step warmup. Training used a batch size of 4096 samples. \\

\textbf{Sparsity and Losses.}
Two SAE variants were trained to minimize the MSE reconstruction loss while enforcing sparsity through different mechanisms: (1)~Top-K SAEs, which enforced fixed sparsity levels (\(k \in \{64,128,256\}\)) via masking, and (2)~Vanilla SAEs, which used \(\ell_1\) regularization with the coefficient swept over \([10^{-12}, 1]\), to induce variable sparsity. To prevent dead features that rarely activate, ghost grads auxiliary loss was used. \\

\textbf{Training Data.}
General-purpose SAEs were trained on ImageNet1k for 1 epoch. For task-specific SAEs, we used CelebA and an augmented Waterbirds dataset, both trained for 2 epochs. The Waterbirds dataset augmentations included rotations (\(\pm10^\circ\), \(\pm20^\circ\)), 5–15\% edge cropping, contrast enhancements (factors of 1.1–1.3), and horizontal flips. \\

\section{Steering Metric}
\label{app:steering}

\subsection{More steering curves}
\label{app:more_steering_curves}

\FloatBarrier

\subsection{Limitations of the steerability metric}\label{app:limitations_steerability}
To address the limitation of the steerability metric not capturing the relationship between semantic concepts, we propose measuring the weighted distance between promoted concepts and 
the mean vocabulary vector:

\begin{equation}
    \mu_V = \frac{1}{|V|}\sum_{v \in V} v
\end{equation}

For a feature $f$, we compute:

\begin{equation}
    D_f = \sum_{v \in V} P(v|f) \|v - \mu_V\|_2
\end{equation}

where $P(v|f)$ is the probability assigned to vocabulary element $v$ by feature $f$. This metric 
shows positive correlation with qualitatively observed monosemantic/steerable features. 

The metric is theoretically justified when vocabulary elements are uniformly distributed on the unit 
sphere (implying $\mu_V \approx 0$), as semantically related concepts cluster together while unrelated 
concepts are approximately orthogonal. We leave the rest to future work.

\onecolumn

\section{Maximally Activating Images}
\label{app:more_max_images}

\begin{figure}[h]
    \centering
    \includegraphics[width=\linewidth]{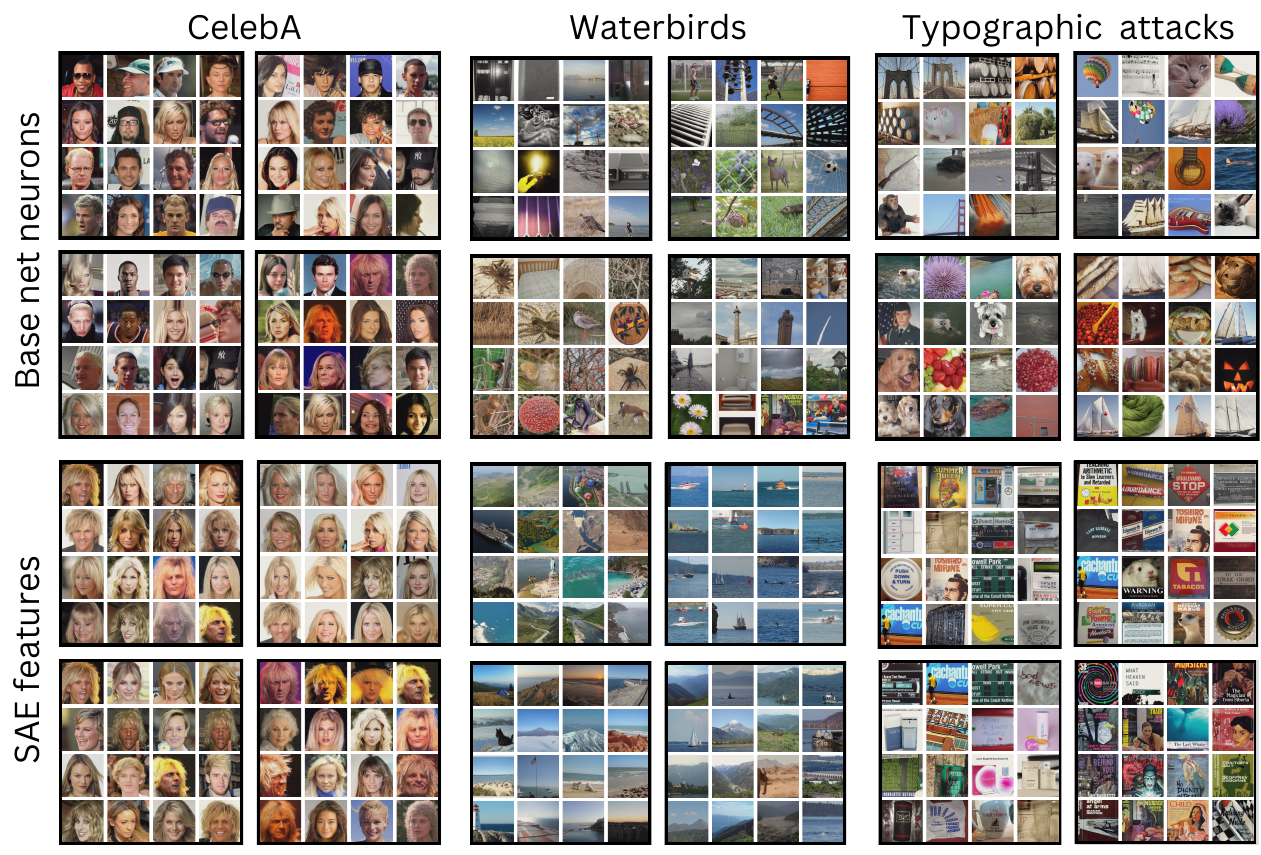}
    \caption{The top activating images show that base net features are polysemantic, while SAE features capture task-relevant attributes: blondeness (CelebA), land/water backgrounds (Waterbirds), and typographic images (typographic attacks).}
    \label{fig:FULL_SAE_neuron_max_act_images}
\end{figure}

\subsection{SAE Evaluations}
\subsubsection{Vanilla SAEs (all patches)}

\begin{table}[H]
\footnotesize 
\setlength{\tabcolsep}{5pt} 
\centering
\begin{tabular}{|l|c|c|c|c|c|c|c|c|c|c|c|c|}
\hline
{\scriptsize\textbf{L1 Coef.}} & 
{\scriptsize\textbf{Exp. Var.}} & 
{\scriptsize\textbf{L0}} & 
{\scriptsize\textbf{Layer}} & 
{\scriptsize\textbf{Sublayer}} & 
{\scriptsize\textbf{Avg Img L0}} & 
{\scriptsize\textbf{Avg CLS L0}} & 
{\scriptsize\textbf{Cos Sim}} & 
{\scriptsize\textbf{Recon Cos Sim}} & 
{\scriptsize\textbf{CE}} & 
{\scriptsize\textbf{Recon CE}} & 
{\scriptsize\textbf{Zero Abl CE}} & 
{\scriptsize\textbf{CE Rec}} \\
\hline
1e-4 & 0.892 & 57.31 & 0 & mlp\_out & 2862.89 & 3.39 & 0.954 & 0.978 & 3.412 & 3.501 & 4.339 & 90.35 \\
8e-5 & 0.910 & 84.88 & 0 & mlp\_out & 4094.50 & 5.13 & 0.962 & 0.982 & 3.415 & 3.491 & 4.342 & 91.77 \\
5e-5 & 0.943 & 160.48 & 0 & mlp\_out & 7718.12 & 8.24 & 0.974 & 0.988 & 3.411 & 3.467 & 4.336 & 93.93 \\
1e-5 & 0.987 & 598.23 & 0 & mlp\_out & 29296.53 & 36.43 & 0.994 & 0.998 & 3.414 & 3.430 & 4.341 & 98.37 \\
\hline
1e-4 & 0.817 & 249.88 & 1 & mlp\_out & 11700.90 & 3.69 & 0.910 & 0.897 & 3.414 & 4.521 & 16.305 & 91.42 \\
8e-5 & 0.851 & 329.70 & 1 & mlp\_out & 15910.30 & 5.28 & 0.928 & 0.921 & 3.417 & 4.252 & 16.309 & 93.52 \\
5e-5 & 0.907 & 592.62 & 1 & mlp\_out & 28769.50 & 6.85 & 0.955 & 0.953 & 3.413 & 3.895 & 16.308 & 96.27 \\
1e-5 & 0.984 & 1478.89 & 1 & mlp\_out & 72397.80 & 88.58 & 0.992 & 0.994 & 3.415 & 3.477 & 16.309 & 99.52 \\
\hline
8e-5 & 0.850 & 390.00 & 2 & mlp\_out & 19407.09 & 13.54 & 0.936 & 0.984 & 3.413 & 3.504 & 6.179 & 96.71 \\
1e-4 & 0.817 & 294.43 & 2 & mlp\_out & 14676.46 & 12.23 & 0.922 & 0.979 & 3.411 & 3.541 & 6.176 & 95.31 \\
5e-5 & 0.903 & 690.89 & 2 & mlp\_out & 33472.50 & 21.24 & 0.958 & 0.991 & 3.412 & 3.460 & 6.180 & 98.28 \\
1e-5 & 0.981 & 1939.23 & 2 & mlp\_out & 88634.21 & 364.95 & 0.992 & 0.999 & 3.413 & 3.421 & 6.183 & 99.71 \\
\hline
1e-4 & 0.777 & 417.00 & 3 & mlp\_out & 20666.06 & 7.29 & 0.905 & 0.988 & 3.415 & 3.497 & 4.581 & 92.93 \\
8e-5 & 0.823 & 572.56 & 3 & mlp\_out & 27773.92 & 8.99 & 0.924 & 0.991 & 3.413 & 3.479 & 4.582 & 94.37 \\
5e-5 & 0.886 & 972.85 & 3 & mlp\_out & 48126.45 & 16.30 & 0.952 & 0.995 & 3.413 & 3.451 & 4.582 & 96.82 \\
1e-5 & 0.982 & 1977.63 & 3 & mlp\_out & 93453.44 & 628.76 & 0.993 & 0.999 & 3.415 & 3.423 & 4.583 & 99.36 \\
\hline
1e-4 & 0.771 & 406.16 & 4 & mlp\_out & 20074.38 & 16.24 & 0.903 & 0.990 & 3.409 & 3.481 & 4.789 & 94.84 \\
8e-5 & 0.816 & 560.38 & 4 & mlp\_out & 27463.11 & 20.32 & 0.923 & 0.992 & 3.411 & 3.465 & 4.788 & 96.09 \\
5e-5 & 0.883 & 984.73 & 4 & mlp\_out & 48444.25 & 35.74 & 0.951 & 0.995 & 3.409 & 3.440 & 4.786 & 97.73 \\
1e-5 & 0.982 & 1975.18 & 4 & mlp\_out & 94364.62 & 1113.69 & 0.993 & 0.999 & 3.412 & 3.416 & 4.789 & 99.72 \\
\hline
1e-4 & 0.768 & 368.31 & 5 & mlp\_out & 18387.25 & 31.38 & 0.903 & 0.990 & 3.409 & 3.470 & 5.142 & 96.46 \\
8e-5 & 0.810 & 506.74 & 5 & mlp\_out & 25240.54 & 39.93 & 0.921 & 0.992 & 3.412 & 3.461 & 5.144 & 97.22 \\
5e-5 & 0.882 & 930.02 & 5 & mlp\_out & 46145.45 & 93.39 & 0.951 & 0.995 & 3.414 & 3.438 & 5.145 & 98.58 \\
1e-5 & 0.983 & 1840.60 & 5 & mlp\_out & 87745.34 & 1284.78 & 0.994 & 0.999 & 3.407 & 3.410 & 5.139 & 99.85 \\

\hline
\end{tabular}
\caption{CLIP-ViT-B-32 vanilla sparse autoencoder performance metrics for all patches.}
\end{table}

\subsubsection{Top-K SAEs (all patches)}

\begin{table}[H]
\footnotesize 
\setlength{\tabcolsep}{5pt} 
\centering
\begin{tabular}{|c|c|c|c|c|c|c|c|c|c|}
\hline
{\scriptsize\textbf{Exp. Var.}} & 
{\scriptsize\textbf{L0}} & 
{\scriptsize\textbf{Layer}} & 
{\scriptsize\textbf{Sublayer}} & 
{\scriptsize\textbf{Cos Sim}} & 
{\scriptsize\textbf{Recon Cos Sim}} & 
{\scriptsize\textbf{CE}} & 
{\scriptsize\textbf{Recon CE}} & 
{\scriptsize\textbf{Zero Abl CE}} & 
{\scriptsize\textbf{CE Rec}} \\
\hline

0.83 & 64 & 1 & resid\_post & 0.888 & 1.0 & 6.762 & 6.762 & 6.908 & 100.0 \\ \hline
0.68 & 64 & 2 & resid\_post & 0.779 & 1.0 & 6.762 & 6.762 & 6.908 & 100.0 \\ \hline
0.70 & 64 & 3 & resid\_post & 0.790 & 1.0 & 6.762 & 6.762 & 6.908 & 100.0 \\ \hline
0.80 & 64 & 4 & resid\_post & 0.858 & 1.0 & 6.762 & 6.762 & 6.908 & 100.0 \\ \hline
0.69 & 64 & 5 & resid\_post & 0.791 & 1.0 & 6.762 & 6.762 & 6.908 & 100.0 \\ \hline
0.78 & 64 & 6 & resid\_post & 0.863 & 1.0 & 6.762 & 6.762 & 6.908 & 100.0 \\ \hline
0.79 & 64 & 7 & resid\_post & 0.881 & 1.0 & 6.762 & 6.762 & 6.908 & 100.0 \\ \hline
0.81 & 64 & 8 & resid\_post & 0.897 & 1.0 & 6.762 & 6.762 & 6.908 & 100.0 \\ \hline
0.83 & 64 & 9 & resid\_post & 0.906 & 1.0 & 6.762 & 6.762 & 6.908 & 100.0 \\ \hline
0.82 & 64 & 10 & resid\_post& 0.893 & 1.0 & 6.762 & 6.762 & 6.908 & 100.0 \\ \hline
0.78 & 64 & 11 & resid\_post& 0.883 & 1.0 & 6.762 & 6.762 & 6.908 & 100.0 \\ \hline

\end{tabular}
\caption{CLIP-ViT-B-32 Top-K sparse autoencoder performance metrics for all patches.}
\end{table}

\section{Disentanglement Task Full Results}
\label{app:disentanglement_tasks}
\begin{table}[htbp]
\caption{Layer-wise Accuracies (\%) for CelebA, under the strict condition}
\small
\begin{tabular}{l*{10}{r}}
\toprule
& \multicolumn{2}{c}{Vanilla} & \multicolumn{2}{c}{Top K=64} & \multicolumn{2}{c}{Top K=128} & \multicolumn{2}{c}{CelebA K=64} & \multicolumn{2}{c}{CelebA K=128}  \\
\cmidrule(lr){2-3} \cmidrule(lr){4-5} \cmidrule(lr){6-7} \cmidrule(lr){8-9} \cmidrule(lr){10-11} 
Layer & Overall & Worst & Overall & Worst & Overall & Worst & Overall & Worst & Overall & Worst \\
\midrule
\multicolumn{11}{c}{SAE Feature Ablation} \\
\midrule
L0 & 92.78 & \textbf{78.89} & 92.78 & 77.78 & \textbf{93.19} & \textbf{79.44} & 92.78 & 77.78 & 92.78 & 77.78 \\
L1 & \textbf{92.92} & 77.78 & 92.78 & 77.78 & \textbf{92.92} & \textbf{78.89} & 92.78 & 77.78 & 92.78 & 77.78 \\
L2 & 92.78 & 77.78 & \textbf{93.47} & \textbf{79.44} & 92.64 & \textbf{78.33} & 92.78 & 77.78 & - & - \\
L3 & 92.64 & \textbf{78.33} & 92.78 & 77.78 & \textbf{93.06} & \textbf{78.89} & 92.78 & 77.78 & 92.78 & 77.78 \\
L4 & \textbf{92.92} & \textbf{78.33} & 92.78 & 77.78 & \textbf{93.19} & \textbf{80.0} & 92.78 & 77.78 & 92.78 & 77.78 \\
L5 & \textbf{93.06} & \textbf{78.89} & \textbf{92.92} & \textbf{78.33} & \textbf{93.19} & \textbf{78.89} & \textbf{93.19} & \textbf{78.89} & 92.64 & \textbf{78.33} \\
L6 & \textbf{93.19} & \textbf{79.44} & \textbf{93.19} & \textbf{79.44} & \textbf{93.06} & \textbf{79.44} & \textbf{93.06} & \textbf{79.44} & \textbf{93.19} & \textbf{80.0} \\
L7* & \textbf{93.06} & \textbf{80.56} & \textbf{93.47} & \textbf{80.56} & \textbf{93.33} & *\textbf{81.11} & \textbf{93.33} & \textbf{80.56} & \textbf{93.61} & *\textbf{81.11} \\
L8* & \textbf{93.61} & \textbf{80.56} & \textbf{93.47} & *\textbf{81.11} & \textbf{93.33} & \textbf{80.56} & \textbf{93.19} & \textbf{80.0} & \textbf{92.92} & \textbf{78.33} \\
L9 & \textbf{92.92} & \textbf{78.89} & \textbf{92.92} & \textbf{78.33} & \textbf{92.92} & \textbf{78.33} & \textbf{93.06} & \textbf{79.44} & - & - \\
L10 & \textbf{93.47} & \textbf{80.0} & \textbf{92.92} & \textbf{78.89} & 92.78 & 77.78 & \textbf{92.92} & \textbf{79.44} & 92.78 & 77.78 \\
L11 & \textbf{93.33} & \textbf{80.56} & 92.78 & 77.78 & \textbf{92.92} & \textbf{78.33} & \textbf{92.92} & \textbf{78.89} & \textbf{93.19} & \textbf{78.89} \\
\midrule
\multicolumn{11}{c}{Random SAE Feature Ablation} \\
\midrule
L0 & 92.78 & 77.78 & 92.78 & 77.78 & 92.78 & 77.78 & 92.78 & 77.78 & 92.78 & 77.78 \\
L1 & 92.64 & 77.78 & 92.78 & 77.78 & 92.78 & 77.78 & 92.78 & 77.78 & 92.78 & 77.78 \\
L2 & 92.64 & 77.78 & 92.78 & 77.78 & 92.78 & 77.78 & 92.78 & 77.78 & - & - \\
L3 & 92.64 & 77.78 & 92.78 & 77.78 & 92.78 & 77.78 & 92.78 & 77.78 & 92.78 & 77.78 \\
L4 & 92.64 & 77.78 & 92.78 & 77.78 & 92.78 & 77.78 & 92.78 & 77.78 & 92.78 & 77.78 \\
L5 & 92.78 & 77.78 & 92.78 & 77.78 & 92.78 & 77.78 & 92.78 & 77.78 & 92.78 & 77.78 \\
L6 & 92.78 & 77.78 & 92.78 & 77.78 & 92.64 & 77.78 & 92.64 & 77.78 & 92.78 & 77.78 \\
L7 & 92.78 & 77.78 & 92.78 & 77.78 & 92.78 & 77.78 & 92.64 & 77.78 & 92.78 & 77.78 \\
L8 & 92.64 & 77.78 & 92.78 & 77.78 & 92.78 & 77.78 & 92.78 & 77.78 & 92.78 & 77.78 \\
L9 & 92.78 & 77.78 & 92.78 & 77.78 & 92.78 & 77.78 & 92.78 & 77.78 & - & - \\
L10 & 92.64 & 77.78 & 92.78 & 77.78 & 92.78 & 77.78 & 92.78 & 77.78 & 92.78 & 77.78 \\
L11 & 92.78 & 77.78 & 92.78 & 77.78 & 92.78 & 77.78 & 92.78 & 77.78 & 92.78 & 77.78 \\
\midrule
\multicolumn{11}{c}{Base Network Neuron Ablation} \\
\midrule
L0 & 92.78 & \textbf{78.89} & - & - & - & - & - & - & - & - \\
L1 & 92.78 & 77.78 & - & - & - & - & - & - & - & - \\
L2 & 92.78 & 77.78 & - & - & - & - & - & - & - & - \\
L3 & 92.78 & 77.78 & - & - & - & - & - & - & - & - \\
L4 & 92.78 & 77.78 & - & - & - & - & - & - & - & - \\
L5 & 92.78 & 77.78 & - & - & - & - & - & - & - & - \\
L6 & 92.78 & 77.78 & - & - & - & - & - & - & - & - \\
L7 & 92.64 & 77.78 & - & - & - & - & - & - & - & - \\
L8 & 92.64 & 77.78 & - & - & - & - & - & - & - & - \\
L9 & 92.78 & 77.78 & - & - & - & - & - & - & - & - \\
L10 & 92.78 & 77.78 & - & - & - & - & - & - & - & - \\
L11 & 92.5 & 77.22 & - & - & - & - & - & - & - & - \\
\bottomrule
\multicolumn{11}{p{\textwidth}}{

\textbf{Note:} Results on the CelebA disentanglement task with targeted ablation of SAE features on the residual stream. The baseline values are 92.78\% for overall accuracy and 77.78\% for worst group accuracy. Bolded values show an improvement over baseline. The best-performing layers (by worst group accuracy) and their corresponding SAEs are marked with an asterisk. The base network ablation does not depend on SAE type, so the value is the same across all SAE types.} \\
\end{tabular}
\label{table:celeba_total_results_strict}
\end{table}

\begin{table}[h]
\caption{Layer-wise Accuracies (\%) for CelebA, under the relaxed condition}
\small
\begin{tabular}{l*{10}{r}}
\toprule
& \multicolumn{2}{c}{Vanilla} & \multicolumn{2}{c}{Top K=64} & \multicolumn{2}{c}{Top K=128} & \multicolumn{2}{c}{CelebA K=64} & \multicolumn{2}{c}{CelebA K=128}  \\
\cmidrule(lr){2-3} \cmidrule(lr){4-5} \cmidrule(lr){6-7} \cmidrule(lr){8-9} \cmidrule(lr){10-11} 
Layer & Overall & Worst & Overall & Worst & Overall & Worst & Overall & Worst & Overall & Worst \\
\midrule
\multicolumn{11}{c}{SAE Feature Ablation} \\
\midrule
L0 & 92.78 & \textbf{78.89} & 92.08 & 76.11 & \textbf{93.61} & \textbf{81.67} & \textbf{92.92} & \textbf{78.33} & \textbf{93.06} & \textbf{78.33} \\
L1 & \textbf{93.06} & \textbf{80.56} & \textbf{93.33} & \textbf{79.44} & \textbf{93.89} & \textbf{80.0} & \textbf{93.33} & \textbf{80.56} & 92.78 & 77.78 \\
L2 & \textbf{93.89} & \textbf{82.78} & \textbf{92.92} & \textbf{81.11} & \textbf{93.47} & \textbf{79.44} & \textbf{94.17} & \textbf{82.22} & - & - \\
L3 & \textbf{94.03} & \textbf{83.33} & \textbf{93.19} & \textbf{81.11} & \textbf{93.06} & \textbf{78.89} & \textbf{93.75} & \textbf{80.56} & \textbf{93.47} & \textbf{82.22} \\
L4 & \textbf{93.33} & \textbf{79.44} & \textbf{93.06} & \textbf{78.33} & \textbf{93.19} & \textbf{80.0} & \textbf{93.19} & \textbf{80.56} & \textbf{93.47} & \textbf{79.44} \\
L5 & \textbf{93.61} & \textbf{80.56} & \textbf{93.47} & \textbf{80.0} & \textbf{93.06} & \textbf{79.44} & \textbf{93.19} & \textbf{78.89} & 92.64 & \textbf{78.33} \\
L6 & \textbf{93.47} & \textbf{81.11} & 92.5 & \textbf{81.11} & \textbf{93.06} & \textbf{79.44} & 92.36 & 77.78 & \textbf{93.19} & \textbf{80.0} \\
L7 & \textbf{93.06} & \textbf{80.56} & \textbf{93.75} & \textbf{84.44} & \textbf{93.47} & \textbf{81.67} & \textbf{94.03} & \textbf{84.44} & \textbf{94.03} & \textbf{84.44} \\
L8 & \textbf{94.03} & \textbf{82.78} & \textbf{93.47} & \textbf{81.11} & \textbf{94.03} & \textbf{85.0} & \textbf{93.89} & \textbf{83.89} & \textbf{94.17} & \textbf{85.56} \\
L9* & \textbf{94.72} & *\textbf{86.67} & \textbf{93.06} & \textbf{79.44} & \textbf{93.47} & \textbf{81.11} & \textbf{93.89} & \textbf{86.11} & - & - \\
L10 & \textbf{93.47} & \textbf{80.0} & \textbf{92.92} & \textbf{78.89} & 92.78 & 77.78 & \textbf{93.19} & \textbf{79.44} & \textbf{94.17} & \textbf{83.89} \\
L11 & \textbf{94.31} & \textbf{83.33} & 92.78 & 77.78 & 92.36 & \textbf{78.33} & \textbf{92.92} & \textbf{78.89} & \textbf{93.19} & \textbf{79.44} \\
\midrule
\multicolumn{11}{c}{Random SAE Feature Ablation} \\
\midrule
L0 & 92.78 & 77.78 & 92.78 & 77.78 & \textbf{92.92} & \textbf{78.33} & 92.78 & 77.78 & 92.64 & 77.78 \\
L1 & 92.64 & 77.78 & 92.78 & 77.78 & 92.78 & 77.78 & 92.78 & 77.78 & 92.78 & 77.78 \\
L2 & 92.64 & 77.78 & 92.78 & 77.78 & 92.78 & 77.78 & 92.5 & \textbf{78.33} & - & - \\
L3 & 92.5 & 77.22 & 92.78 & 77.78 & 92.78 & 77.78 & 92.78 & 77.78 & 92.78 & 77.78 \\
L4 & 92.78 & 77.78 & 92.78 & 77.78 & 92.78 & 77.78 & 92.78 & 77.78 & 92.78 & 77.78 \\
L5 & 92.64 & 77.78 & 92.64 & 77.78 & 92.64 & 77.78 & 92.78 & 77.78 & 92.78 & 77.78 \\
L6 & 92.78 & \textbf{78.89} & \textbf{92.92} & 77.78 & 92.64 & 77.78 & 92.78 & 77.78 & 92.78 & 77.78 \\
L7 & 92.78 & 77.78 & 92.64 & 77.78 & 92.5 & 77.22 & 92.64 & 77.78 & 92.64 & 77.78 \\
L8 & 92.78 & \textbf{78.33} & 92.78 & 77.78 & 92.64 & 77.22 & 92.64 & 77.78 & \textbf{92.92} & \textbf{78.33} \\
L9 & 92.64 & 77.78 & 92.78 & 77.78 & 92.64 & 77.78 & \textbf{92.92} & \textbf{78.89} & - & - \\
L10 & 92.64 & 77.78 & 92.78 & 77.78 & 92.78 & 77.78 & 92.78 & 77.78 & 92.64 & 77.78 \\
L11 & 92.78 & 77.78 & 92.78 & 77.78 & 92.78 & 77.78 & 92.78 & 77.78 & 92.5 & 77.22 \\
\midrule
\multicolumn{11}{c}{Base Network Neuron Ablation} \\
\midrule
L0 & \textbf{93.19} & \textbf{80.0} & - & - & - & - & - & - & - & - \\
L1 & \textbf{93.47} & \textbf{80.56} & - & - & - & - & - & - & - & - \\
L2 & \textbf{93.47} & \textbf{79.44} & - & - & - & - & - & - & - & - \\
L3 & 92.78 & 77.78 & - & - & - & - & - & - & - & - \\
L4 & \textbf{92.92} & 77.78 & - & - & - & - & - & - & - & - \\
L5 & \textbf{92.92} & 77.78 & - & - & - & - & - & - & - & - \\
L6 & 92.78 & 77.78 & - & - & - & - & - & - & - & - \\
L7 & \textbf{93.75} & \textbf{80.0} & - & - & - & - & - & - & - & - \\
L8 & 92.64 & 77.78 & - & - & - & - & - & - & - & - \\
L9 & 92.78 & 77.78 & - & - & - & - & - & - & - & - \\
L10 & 92.78 & 77.78 & - & - & - & - & - & - & - & - \\
L11 & \textbf{93.33} & \textbf{79.44} & - & - & - & - & - & - & - & - \\
\bottomrule
\multicolumn{11}{p{\linewidth}}{
\textbf{Note:} The baseline values are 92.78\% for overall accuracy and 77.78\% for worst group accuracy. Base network ablation is not dependent on the SAE type, so the value is the same across all SAE types.} \\
\end{tabular}
\label{table:celeba_total_results_relaxed}
\end{table}

\begin{table}[ht]
\caption{Layer-wise Accuracies (\%) for Waterbirds, under the strict condition}
\small
\begin{tabular}{l*{10}{r}}
\toprule
& \multicolumn{2}{c}{Vanilla} & \multicolumn{2}{c}{Top K=64} & \multicolumn{2}{c}{Waterbirds K=64}  \\
\cmidrule(lr){2-3} \cmidrule(lr){4-5} \cmidrule(lr){6-7} 
Layer & Overall & Worst & Overall & Worst & Overall & Worst \\
\midrule
\multicolumn{11}{c}{SAE Feature Ablation} \\
\midrule
L0 & 68.81 & 22.43 & \textbf{69.42} & 20.25 & 68.81 & 22.43 \\
L1 & 68.81 & 22.43 & 68.81 & 22.43 & 68.81 & 22.43 \\
L2* & \textbf{68.99} & *\textbf{24.61} & \textbf{71.95} & 21.65 & 68.81 & 22.43 \\
L3 & \textbf{69.54} & \textbf{24.14} & 68.81 & 22.43 & 68.81 & 22.43 \\
L4 & \textbf{68.93} & \textbf{23.21} & 68.81 & 22.43 & 68.81 & 22.43 \\
L5 & \textbf{69.05} & \textbf{22.74} & 68.81 & 22.43 & 68.81 & 22.43 \\
L6 & 68.81 & 22.43 & 68.81 & 22.43 & 68.81 & 22.43 \\
L7 & 68.81 & 22.43 & 68.81 & 22.43 & 68.81 & 22.43 \\
L8 & 68.81 & 22.43 & 68.81 & 22.43 & 68.81 & 22.43 \\
L9 & \textbf{70.19} & \textbf{22.74} & 68.81 & 22.43 & 68.81 & 22.43 \\
L10 & 68.81 & 22.43 & 68.81 & 22.43 & 68.81 & 22.43 \\
L11 & 68.81 & 22.43 & 68.81 & 22.43 & 68.81 & 22.43 \\
\midrule
\multicolumn{11}{c}{Random SAE Feature Ablation} \\
\midrule
L0 & 68.81 & 22.43 & 68.81 & 22.43 & 68.81 & 22.43 \\
L1 & 68.81 & 22.43 & 68.81 & 22.43 & 68.81 & 22.43 \\
L2 & 68.81 & 22.43 & 64.2 & 22.27 & 68.81 & 22.43 \\
L3 & \textbf{68.85} & 22.27 & 68.81 & 22.43 & 68.81 & 22.43 \\
L4 & 68.81 & 22.43 & 68.81 & 22.43 & 68.81 & 22.43 \\
L5 & 68.81 & 22.43 & 68.81 & 22.43 & 68.81 & 22.43 \\
L6 & 68.81 & 22.43 & 68.81 & 22.43 & 68.81 & 22.43 \\
L7 & 68.81 & 22.43 & 68.81 & 22.43 & 68.81 & 22.43 \\
L8 & 68.81 & 22.43 & 68.81 & 22.43 & 68.81 & 22.43 \\
L9 & 68.81 & 22.43 & 68.81 & 22.43 & 68.81 & 22.43 \\
L10 & 68.81 & 22.43 & 68.81 & 22.43 & 68.81 & 22.43 \\
L11 & 68.81 & 22.43 & 68.81 & 22.43 & 68.81 & 22.43 \\
\midrule
\multicolumn{11}{c}{Base Network Neuron Ablation} \\
\midrule
L0 & 68.81 & 22.43 & - & - & - & - \\
L1 & 68.81 & 22.43 & - & - & - & - \\
L2 & 68.81 & 22.43 & - & - & - & - \\
L3 & 68.81 & 22.43 & - & - & - & - \\
L4 & 68.81 & 22.43 & - & - & - & - \\
L5 & 68.81 & 22.43 & - & - & - & - \\
L6 & 68.81 & 22.43 & - & - & - & - \\
L7 & 68.81 & 22.43 & - & - & - & - \\
L8 & 68.81 & 22.43 & - & - & - & - \\
L9 & 68.81 & 22.43 & - & - & - & - \\
L10 & 68.81 & 22.43 & - & - & - & - \\
L11 & 68.81 & 22.43 & - & - & - & - \\
\bottomrule
\multicolumn{11}{p{\linewidth}}{
\textbf{Note:} The baseline values are 68.81\% for overall accuracy and 22.43\% for worst group accuracy.} \\
\end{tabular}
\label{table:waterbirds_total_results_strict}
\end{table}

\begin{table}[t]
\caption{Layer-wise Accuracies (\%) for Waterbirds, under the relaxed condition}
\small
\begin{tabular}{l*{10}{r}}
\toprule
& \multicolumn{2}{c}{Vanilla} & \multicolumn{2}{c}{Top K=64} & \multicolumn{2}{c}{Waterbirds K=64}  \\
\cmidrule(lr){2-3} \cmidrule(lr){4-5} \cmidrule(lr){6-7} 
Layer & Overall & Worst & Overall & Worst & Overall & Worst \\
\midrule
\multicolumn{11}{c}{SAE Feature Ablation} \\
\midrule
L0 & 68.81 & 22.43 & \textbf{69.42} & 20.25 & 68.81 & 22.43 \\
L1 & 68.81 & 22.43 & \textbf{69.68} & \textbf{30.84} & \textbf{70.26} & 19.16 \\
L2 & \textbf{68.99} & \textbf{24.61} & \textbf{71.95} & 21.65 & \textbf{71.94} & \textbf{23.68} \\
L3 & \textbf{68.86} & \textbf{33.96} & 68.81 & 22.43 & \textbf{70.25} & 19.31 \\
L4 & \textbf{70.19} & \textbf{27.73} & \textbf{68.83} & \textbf{27.26} & 68.81 & 22.43 \\
L5* & \textbf{70.07} & *\textbf{36.6} & \textbf{70.76} & \textbf{25.55} & 68.81 & 22.43 \\
L6 & \textbf{70.04} & \textbf{33.18} & \textbf{69.0} & \textbf{24.14} & \textbf{71.37} & \textbf{30.22} \\
L7 & \textbf{70.62} & \textbf{28.66} & 68.81 & 22.43 & 68.81 & 22.43 \\
L8 & \textbf{71.3} & \textbf{24.92} & 68.81 & 22.43 & 68.81 & 22.43 \\
L9 & \textbf{70.19} & \textbf{22.74} & 68.81 & 22.43 & 68.66 & 22.43 \\
L10 & 68.73 & 22.27 & 68.81 & 22.43 & 68.81 & 22.43 \\
L11 & \textbf{69.12} & \textbf{23.83} & \textbf{69.68} & \textbf{22.59} & \textbf{70.25} & 21.18 \\
\midrule
\multicolumn{11}{c}{Random SAE Feature Ablation} \\
\midrule
L0 & 68.81 & 22.43 & 68.81 & 22.43 & 68.81 & 22.43 \\
L1 & 68.81 & 22.43 & 68.81 & 22.43 & 68.81 & 22.43 \\
L2 & 68.81 & 22.43 & 64.2 & 22.27 & 68.81 & 22.43 \\
L3 & \textbf{68.83} & 22.43 & 68.81 & 22.43 & 68.78 & 22.43 \\
L4 & 68.8 & 22.12 & 68.81 & 22.43 & 68.81 & 22.43 \\
L5 & 68.62 & 22.43 & \textbf{68.85} & 22.43 & 68.81 & 22.43 \\
L6 & 68.59 & 22.27 & 68.81 & 22.43 & \textbf{68.83} & \textbf{22.59} \\
L7 & \textbf{68.92} & 22.43 & 68.81 & 22.43 & 68.81 & 22.43 \\
L8 & 68.76 & 22.43 & 68.81 & 22.43 & 68.81 & 22.43 \\
L9 & 68.81 & 22.43 & 68.81 & 22.43 & 68.81 & 22.43 \\
L10 & 68.81 & 22.43 & 68.81 & 22.43 & 68.81 & 22.43 \\
L11 & \textbf{68.83} & \textbf{22.59} & 68.81 & 22.43 & 68.8 & 22.43 \\
\midrule
\multicolumn{11}{c}{Base Network Neuron Ablation} \\
\midrule
L0 & 68.81 & 22.43 & - & - & - & - \\
L1 & 68.81 & 22.43 & - & - & - & - \\
L2 & 68.81 & 22.43 & - & - & - & - \\
L3 & 68.81 & 22.43 & - & - & - & - \\
L4 & 68.81 & 22.43 & - & - & - & - \\
L5 & 68.81 & 22.43 & - & - & - & - \\
L6 & 68.81 & 22.43 & - & - & - & - \\
L7 & 68.81 & 22.43 & - & - & - & - \\
L8 & 68.81 & 22.43 & - & - & - & - \\
L9 & 68.81 & 22.43 & - & - & - & - \\
L10 & 68.81 & 22.43 & - & - & - & - \\
L11 & 68.81 & 22.43 & - & - & - & - \\
\bottomrule
\multicolumn{11}{p{\textwidth}}{
\textbf{Note:} The baseline values are 68.81\% for overall accuracy and 22.43\% for worst group accuracy.} \\
\end{tabular}
\label{table:waterbirds_total_results_relaxed}
\end{table}

\end{document}